\documentclass{article}
\usepackage{spconf,amsmath,graphicx}
\usepackage{pgf,tikz,pgfplots}
\usepackage{booktabs}
\usepackage{xcolor}
\usepackage{array} % for "\extrarowheight" macro
\usepackage[skip=0.333\baselineskip]{caption}
\usepackage{amsmath}
\usepackage{amssymb}
\usepackage{amsthm}
\usepackage{mathtools}
\usepackage{stackengine}

\DeclareMathOperator*{\argmin}{arg\,min}

\usepackage{dblfloatfix}
\usepackage{graphicx}
\usepackage{pgf,tikz,pgfplots}
\usetikzlibrary{arrows,shapes}
\usetikzlibrary{plotmarks}
\usepgfplotslibrary{groupplots}
\pgfplotsset{compat=newest}
\pgfplotsset{plot coordinates/math parser=false}
\usetikzlibrary{positioning,spy}
\usetikzlibrary{backgrounds}

\usepackage{comment}
\usepackage{multirow}
\usepackage{tabularx}
\usepackage{colortbl}
\usepackage{xcolor}
\usepackage{soul}
\usepackage[normalem]{ulem}
\usepackage{hyperref}
\title{Deep Coding Patterns Design for Compressive Near-Infrared Spectral Classification} 
\name{Jorge Bacca$^1$, Alejandra Hernandez-Rojas$^2$ and Henry Arguello$^1$\thanks{This material is based upon work supported by the Air Force Office of Scientific Research under award number FA9550-21-1-0326}}
\address{$^1$Department of Computer Science, Universidad Industrial de Santander, Bucaramanga, Colombia\\ $^2$Department of Geophysics,
	Universidad Industrial de Santander, Bucaramanga, Colombia}
\begin{document}
%\ninept
%
\maketitle
\begin{abstract}

Compressive spectral imaging (CSI) has emerged as an attractive compression and sensing technique, primarily to sense spectral regions where traditional systems result in highly costly such as in the near-infrared spectrum. Recently, it has been shown that spectral classification can be performed directly in the compressive domain, considering the amount of spectral information embedded in the measurements, skipping the reconstruction step. Consequently, the classification quality directly depends on the set of coding patterns employed in the sensing step. Therefore, this work proposes an end-to-end approach to jointly design the coding patterns used in CSI and the network parameters to perform spectral classification directly from the embedded near-infrared compressive measurements. Extensive simulation on the three-dimensional coded aperture snapshot spectral imaging (3D-CASSI) system validates that the proposed design outperforms traditional and random design in up to 10\% of classification accuracy.
\end{abstract}
\begin{keywords}
Compressive Spectral Imaging, Near-Infrared Spectral Imaging, Spectral Classification, Deep Coded Design.\vspace{-0.5em}
\end{keywords}
\section{Introduction}\vspace{-1em}
\label{sec:intro}

Hyperspectral imaging (HSI) is a technique that combines conventional digital imaging and spectroscopy in a single system. It provides three-dimensional (3D) images, called spectral data cube, which captures spatial information along the electromagnetic spectrum~\cite{Kamruzzaman2011}. While substantial progress has been made in the visible spectral region, up to 1100 nm, the progress in integrating of near-infrared (NIR) spectrum has been limited due to the optical material costs \cite{Hakkel2022}. The NIR part of the electromagnetic spectrum, particularly the region 1000-1700 nm, provides a lot of information to determine the quality and safety of agricultural and food products \cite{Elmasry2012}. Other fields of interest and research areas where NIR hyperspectral imaging is applied include pharmaceuticals \cite{Otsuka2020}, medical applications \cite{Sakudo2016}, archaeology \cite{Linderholm2019}, and paleontology \cite{Thomas2011} since it provides higher sensitivity than solutions in the visible spectrum~\cite{Hakkel2022}.

Conventional HSI scanning sensing approaches require high storage capacity and computational costs, as large exposure time to scan each spectral voxel \cite{arce2014compressive}. Furthermore, a preprocessing step to reduce the dimensions of the spectral imagery is often required \cite{Rasti2016}. Compressive spectral imaging (CSI) has emerged as a spectral imaging acquisition approach that captures 2D compressive measurements of the entire data cube rather than directly acquiring all the voxels, thus reducing the data dimensionality \cite{arce2014compressive}. For instance, the 3D-coded aperture snapshot spectral imaging (3D-CASSI), naturally embodies the principles of compressive sensing by acquiring the entire spectral data cube with just focal plane array (FPA) measurements employing binary coding patterns~\cite{arce2014compressive}. Subsequently, assuming a prior knowledge of the scene as sparsity, global spectral correlation, or non-local self-similarity priors \cite{arce2014compressive, Chen2020}, the spectral image can be recovered from the compressive measurements using optimization-based, iterative algorithms or more recently, deep learning approaches~\cite{mousavi2017learning, Jin2017}. However, these approaches require expensive computational costs for the signal recovery from its compressed measurements \cite{kulkarni2016reconnet, bacca2019noniterative}.

%Spectral classification has attracted much attention because it can be used to learning patterns to discretize samples in a spectral data cube.
Recently, compressed learning has demonstrated that, for inference tasks such as segmentation, detection, or classification, the reconstruction can be avoided by solving the problem directly in the compressive domain \cite{hinojosa2018coded, Hinojosa2019, ramirez2014spectral, lohit2016direct}. For instance, compressive spectral classification has been used in agriculture and remote sensing applications since it allows identifying materials in the scene using only the compressive spectral signature. ~\cite{hinojosa2018coded,Elmasry2012}.
The performance of the inference task directly depends on the coding pattern structure, which is usually a binary sensing matrix. For instance, in compressive spectral classification, \cite{ramirez2014spectral} proposed to employ random binary coding patterns, and \cite{hinojosa2018coded} proposed a coding patterns design based on a greedy algorithm that preserves the similarity in the visible electromagnetic range among the spectral signatures. On the other hand, works in  \cite{bacca2020coupled, bacca2021deep} used a deep learning approach to simultaneously learn the binary pattern in a single-pixel camera (SPC) using a non-linear classification net. 

In contrast to state-of-the-art methods, this work proposes an end-to-end (E2E) model that consists of an optimization framework to jointly learn the binary coding patterns and the optimal parameters of a  non-linear NIR spectral classification network. To employed the E2E framework we need to model the first layer of the network as a fully differentiable optical layer to simulate the compressive measurements, whereas the subsequent layers learn the inference task. After that, the binary coding patterns can be used in a real 3D-CASSI system to acquire compressed measurements that can be classified using the trained subsequent part of the network. To the best of our knowledge, this is the first work to design binary coding patterns using deep learning methods for spectral classification using the infrared-spectral range. The performance of the proposed approach is demonstrated on two hyperspectral datasets in the NIR region: Cereals \cite{gowen2019comparison} in the spectral range of 943-1643 nm, and Yatsuhashi, which covers 1293-2215 nm. The results show better average classification accuracy above $90\%$ for different sensing ratios compared to coding strategies based on the visible range \cite{hinojosa2018coded} and \cite{xu2020deep}, specifically, when the sensing ratio decreases, where the reached difference is up to $10 \%$ of accuracy.

\section{Compressive measurements acquisition}

\begin{figure}[!t]
    \centering
    \includegraphics[width=1\columnwidth]{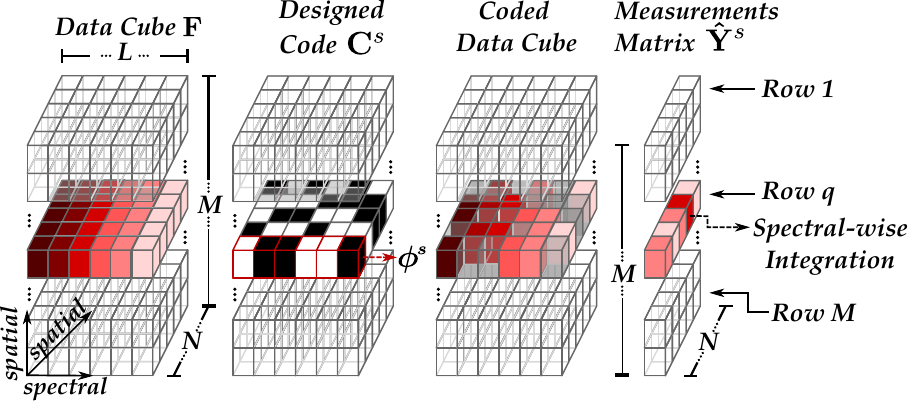}
    \caption{3D-CASSI sensing approach at the snapshot $s$.}
    \label{fig:3D_CASSI}
\end{figure}

The 3D-coded aperture snapshot spectral imaging (3D-CASSI) is a CSI sensing scheme that allows a spatio-spectral modulation on the spectral data cube using different coding patterns for each spectral band \cite{cao2016computational} as illustrated in Fig.\ref{fig:3D_CASSI}. Let $\mathcal{F} \in \mathbb{R}^{M \times N \times L}$ be the spatio-spectral input data cube with $M \times N$ spatial dimensions and $L$ spectral bands. Each voxel of an $M \times N \times L$ spectral image is denoted as $f_{m,n,l}$, where $m=0,...,M-1$ and $n=0,\ldots,N-1$ index the spatial coordinates and $l=0,\ldots,L-1$ represents the spectral band index. Specifically, each spectral pixel passes through its corresponding binary coding pattern indexed as $\phi^s_{m,n,l}\in \{0,1\}$, where $s=0,...,S-1$ index the number of snapshots. Then, the coded spectral scene is relayed into the focal plane array (FPA) detector, where the compressive measurements ($S\ll L$) are acquired by the integration over the detector's spectral dimension. Mathematically, the output of the sensing process at the ($m,n$)-th detector pixel and a specific snapshot $s$ can be expressed as:
\begin{equation}
    \hat{y}^s_{m,n}=\sum_{l=0}^{L-1}\phi^s_{m,n,l}f_{m,n,l},
\label{eq:sensing}
\end{equation}
The set of compressive measurements from (\ref{eq:sensing}) can be arranged in a $S \times MN$ matrix $\mathbf{\hat{Y}}$, where each column contains the compressive measurements associated with a particular spectral pixel. In practice, the number of coding patterns is less than the number of pixels. Therefore, this work assumes that the number of different coding patterns in a single shot is equal to $S$, and to reduce the redundant information, each pixel is encoded only once by a different coding pattern, i.e., at the end of the sensing procedure, all pixels are encoded by the whole set of $S$ coding patterns~\cite{hinojosa2018coded}. Consequently, the entries of $\mathbf{\hat{Y}}$ can be rearranged to form a new matrix $\mathbf{Y}$, such that each row contains the compressed information acquired with a specific coding pattern $\phi^s$. Formally, the rearrangement can be expressed as:

\begin{equation}
    Y_{s,j}=\hat{Y}_{s',j}~ ~\mathrm{if}~ ~\hat{Y}_{s',j}=(\phi^s)^T\mathbf{f}_j~ ~\forall s',
    \label{eq:rearrangement}
\end{equation}
for $s,s'=0,\ldots,S-1$, where $\mathbf{f}_j$ denotes the $j$-th spectral signature with $j=0,\ldots,MN-1$. This rearrangement \cite{hinojosa2018coded}, preserves the structure of the high-dimensional data. Alternatively, defining the matrix of $S$ coding patterns as $\mathbf{H}=[\phi^0,\phi^1, \ldots, \phi^{S-1}]^T$, the sensing problem can be simplified as:
\begin{equation}
    \mathbf{Y=HF},
\end{equation}
where $\mathbf{F} \in \mathbb{R}^{L\times MN}$ is the spatio-spectral data cube in matrix form, and $\mathbf{H} \in \lbrace0,1\rbrace^{S\times L}$ the projection matrix, where each row represents a coding pattern. A conventional relation that defines the compression achieved by these systems is the sensing ratio, calculated as $\%_{3D}= \dfrac{S}{L}$.

A common procedure after the compressed measurements are acquired is the signal recovery, which is achieved using nonlinear and relatively expensive optimization-based or iterative algorithms \cite{figueiredo2007gradient}. However, for some specific tasks, such as segmentation, detection, or classification, the reconstruction can be avoided by solving the problem directly in the compressive domain \cite{hinojosa2018coded, bacca2020coupled}. Indeed, \cite{hinojosa2018coded,bacca2020coupled} have shown that the classification accuracy over compressed measurements can be improved if the set of coded apertures is appropriately designed.

\section{Proposed Deep Coding Patterns Design}

\begin{figure}[!t]
    \centering
    \includegraphics[width=1\columnwidth]{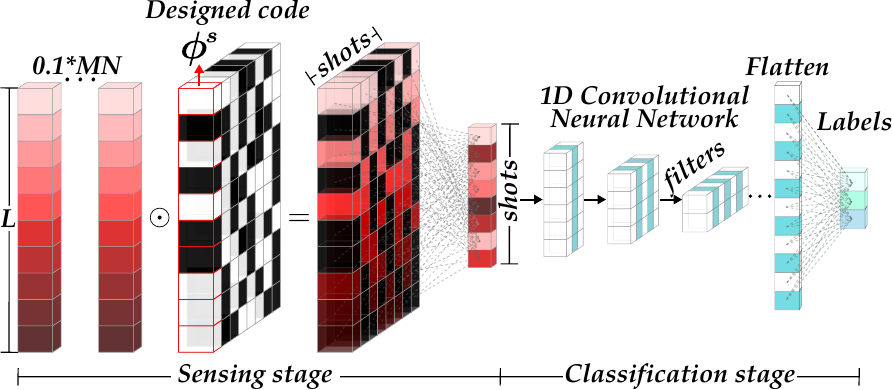}\vspace{1em}
    \caption{Proposed deep learning scheme. In the sensing stage, the designed binary code is learned and implemented to acquire the compressive measurements, which are the input to train the classification neural network.}
    \label{fig:NN_scheme}
\end{figure}

The proposed method consists in modeling the 3D-CASSI system as an optical layer. Considering the rearrange step expressed in \eqref{eq:rearrangement}, the forward sensing model can be treated as a fully-connected layer, i.e., each coding pattern is multiplied element-wise with each spectral signature, where the number of patterns is equivalent to the number of neurons as illustrated in the sensing stage of Fig.\ref{fig:NN_scheme}. Therefore, the optical system can be modeled as a layer:

\begin{equation}
    \mathbf{Y}= \mathbf{HF}:=\mathcal{M}_{\boldsymbol{\phi}}(\mathbf{F}),
\end{equation}
where $\boldsymbol{\phi}$ denotes the coding patterns, which are the learnable parameters in the fully-connected model  $\mathcal{M}_{\boldsymbol{\phi}}$. Notice that the main difference with the traditional fully-connected layer is that the entries of the coding patterns need to be binary. The following section shows how to address the binary constraint from the E2E optimization design~\cite{arguello2021shift}.
\subsection{End-to-End Optimization}
Notice that the sensing measurements $\mathbf{Y}$ can be used as a feature matrix extracted from $\mathbf{F}$, and consequently, it is possible to perform spectral classification directly to the embedded measurements. Therefore, this work uses $\mathcal{N}_{\theta}$ as a deep model to classify the compressive measurements, where $\theta$ denotes the learnable parameters. To learn the coding patterns and the parameters of the deep model, this work is based on the E2E optimization, which is mathematically expressed as:
\begin{equation*}
\hspace{-0em}\{\boldsymbol{\phi}^*,\boldsymbol{\theta}^*\} \hspace{-0em}=\hspace{-0em}\argmin_{\boldsymbol{\phi}, \boldsymbol{\theta}} \sum_{k}\hspace{-0em} \mathcal{L}_
		{loss}\left(\mathcal{N}_{\boldsymbol{\theta}}\left(\mathcal{M}_{\boldsymbol{\phi}}(\mathbf{f}_k )\right) , \mathbf{d}_k\right)\hspace{-0em}+\hspace{-0em}\rho R(\boldsymbol{\phi}),
	\end{equation*}
where $\{\boldsymbol{\phi}^*,\boldsymbol{\theta}^*\}$ represent the optimal optical coding parameters and the optimal weights of the network, respectively; $\{\mathbf{f}_{k},\mathbf{d}_k\}_{k = 1}^{K}$, account for the training database, with $\mathbf{f}_{k}$ as the spectral signature and $\mathbf{d}_{k}$ as the classification label. The loss function $\mathcal{L}_{loss}$ for the classification task is the categorical cross-entropy. Since the coding patterns need to be binary, this work uses $R(\boldsymbol{\phi})$ as a regularization function that acts only over the coding patterns to promote binary values, with $\rho$ as a control parameter~\cite{bacca2020coupled}. In particular, the regularization function is included in the optimization problem as:
	\begin{equation}
		R_1(\boldsymbol{\phi}) = \frac{1}{n} \sum_{{\ell=1}}^{n} ({\boldsymbol{\phi}_{{\ell}}})^2({\boldsymbol{\phi}_{{\ell}}}-1)^2,
		\label{eq:family_binary_0}
	\end{equation}
which is minimized when the elements of the coding patterns are either $0$ or $1$. The proposed E2E model which simultaneously learns the coding patterns and the parameters of the classification network is summarized in Fig \ref{fig:NN_scheme}.

Since the parameters $\boldsymbol{\phi}$ and $\boldsymbol{\theta}$ can be jointly optimized with efficient stochastic gradient algorithms employed in the training of the neural network, the main idea of including the regularization in the training is that the gradient of the loss function concerning $\boldsymbol{\phi}$ is calculated using the chain rule as:
\begin{equation}
\frac{\partial \mathcal{L}}{\partial \boldsymbol{\phi}} = \frac{\partial \mathcal{L}_{loss}}{\partial \mathcal{N}_{\theta}}\frac{\partial \mathcal{N}_{\theta}}{\partial \mathcal{M}_{\boldsymbol{\phi}}} \frac{\partial \mathcal{M}_{\boldsymbol{\phi}}}{\partial \boldsymbol{\phi}} + \rho\frac{\partial R}{\partial \boldsymbol{\phi}}
\end{equation}

Therefore, the design of the coding patterns is directly influenced by the loss of the classification task and the one given by the physical constraints. Consequently, the parameter $\rho$ plays an essential role in the optimal performance and coded aperture implementability. This work uses an exponential increase strategy~\cite{bacca2021deep,https://doi.org/10.48550/arxiv.2205.12158}, which consist of starting with low values of $\rho$ at the first epochs to obtain the direction of the desired task values, and then $\rho$ is increased to guarantee binary values.

\section{Simulations and Results}
\begin{figure}[!t]
    \centering
    \includegraphics[width=0.9\columnwidth]{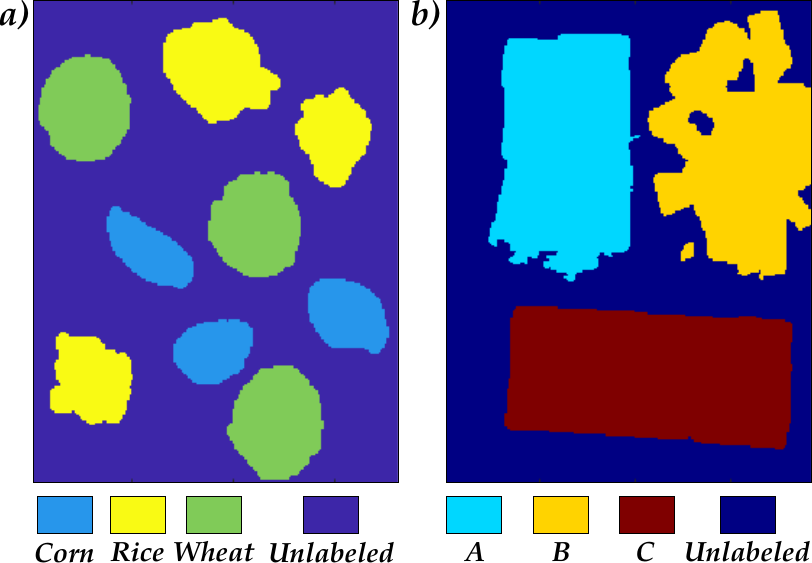}\vspace{1em}
    \caption{Ground truth of the NIR spectral images for a) Cereal Dataset and b) Yatsuhashi Dataset.}
    \label{fig:gt}
\end{figure}
The proposed coding patterns design for classification was tested on two hyperspectral datasets in the NIR region. The first dataset called \textbf{Cereals} contains reflectance of cereal samples in the spectral range of 943-1643 nm with an interval of 7 nm leading to 101 spectral bands~\cite{gowen2019comparison}. The spatial dimension of this image is $241\times 181$ pixels, which included three types of puffed cereals labeled as Corn, Wheat, and Rice, according to their main components. The second NIR dataset, \textbf{Yatsuhashi}\footnote{Available in \href{https://www.kaggle.com/hacarus/near-infrared-hyperspectral-image}
 {Near Infrared Hyperspectral Image Dataset}. Accessed: 20-Jan-2022.} contains $192 \times 256$ pixels with $96$ spectral bands covering 1293-2215 nm and is composed of three types of yatsuhashi sweets commercialized by different companies. The ground truths with the label classes for both datasets are shown in Fig.~\ref{fig:gt}.

The classification network structure used in the E2E model was inspired in the work presented in \cite{xu2020deep} for NIR images. Specifically, the network consists of three convolutional layers with batch normalization and rectified linear unit (ReLU) as the activation function, followed by a dropout to avoid overfitting, and finally, two fully connected layers, where the last one has the softmax activation. This network was trained using $10\%$ of the total spectral signatures. The proposed deep coding patterns design was compared with a random pattern (Random-design), and the coded design proposed in \cite{hinojosa2018coded} was denoted as (Traditional-design). Additionally, the classification results when using the whole spectral data cube (Full-data) is included, i.e., no compression is performed ($\%_{3D}=1$), and the sensing matrix is $\mathbf{H=I}$.
%, the network weights were random initialized
\subsection*{Compression ratio analysis}

\begin{table}[t!]
	\scriptsize
	\renewcommand{\arraystretch}{0.85}\vspace{1em}
	\caption{Quantitative evaluation for the coding patterns design in terms of overall accuracy for the Cereals Dataset}
	\label{tab:Cereals}
	\centering
	\resizebox{1\columnwidth}{!}{%
	\setlength\tabcolsep{0.05cm}
	\begin{tabular}{lcccc} 
		\toprule 
		$\%_{3D}$ & Random-design & Traditional-design & Deep-design & Full-data  \\ \midrule 
	%	VDSR & 4 & 6.7004 &0.0419 & 6.6991 & 27.8821& 0.7242 \\ 
		0.1 & 0.8003 & \underline{0.8856} &\textbf{0.9117} & - \\ 
		0.2 & 0.8663 & \underline{0.9139} & \textbf{0.9275} & - \\ 
		0.3 & 0.8399 & \underline{0.9232} & \textbf{0.9386} & - \\ 
		0.4 & 0.8753 & \underline{0.9359} & \textbf{0.9440} & - \\ 
		0.5 & 0.8772 & \underline{0.9398} & \textbf{0.9596} & - \\
		1 & - & - &-  & \textbf{0.9785} \\ 
		\bottomrule 
	\end{tabular}  }
\end{table}

\begin{table}[b!]
	\scriptsize
	\renewcommand{\arraystretch}{0.85}\vspace{1em}
	\caption{Quantitative evaluation for the coding patterns design in terms of overall accuracy for the Yatsuhashi Dataset}
	\label{tab:Coffe}
	\centering
	\resizebox{1\columnwidth}{!}{%
	\setlength\tabcolsep{0.05cm}
	\begin{tabular}{lcccc} 
		\toprule 
		$\%_{3D}$ & Random-design & Traditional-design & Deep-design & Full-data  \\ \midrule 
	%	VDSR & 4 & 6.7004 &0.0419 & 6.6991 & 27.8821& 0.7242 \\ 
		0.1 & 0.9192 & \underline{0.9322} &\textbf{0.9671} & - \\ 
		0.2 & 0.9328 & \underline{0.9488} &\textbf{0.9681} & - \\ 
		0.3 & 0.9458 &  \underline{0.9483} &\textbf{0.9684} & - \\ 
		0.4 & \underline{0.9542} & 0.9528 &\textbf{0.9708} & - \\ 
		0.5 & 0.9610 & \underline{0.9652} &\textbf{0.9714} & - \\
		1 & - & - &-  & \textbf{0.9894} \\ 
		\bottomrule 
	\end{tabular}  }
\end{table}
 
Numerical tests were conducted to demonstrate the proposed coding patterns design under different sensing ratios $\%_{3D}=[0.1,0.2,0.3,0.4,0.5]$ where $0.1$ is the extreme case of compression evaluated. Tables \ref{tab:Cereals} and \ref{tab:Coffe} summarize the classification accuracy obtained for both datasets, by the selection of the best experiment obtained of a total of $25$ training trials. For all the scenarios, the proposed design outperforms the other designs in up to 10\% of accuracy. Additionally, the main gain is obtained with the highest compression value ($\%_{3D}=0.1$) which is the desired performance in order to sense fewer data. Also, the result employing the full-data is also presented, i.e., without compression ($\%_{3D}=1$), where it can be observed that the compressive classification in the NIR spectrum is possible and the accuracy difference is less than $2\%$ in comparison with the Deep-design for $\%_{3D}=0.5$.

\begin{figure}[!t]
    \centering
    \includegraphics[width=1\columnwidth]{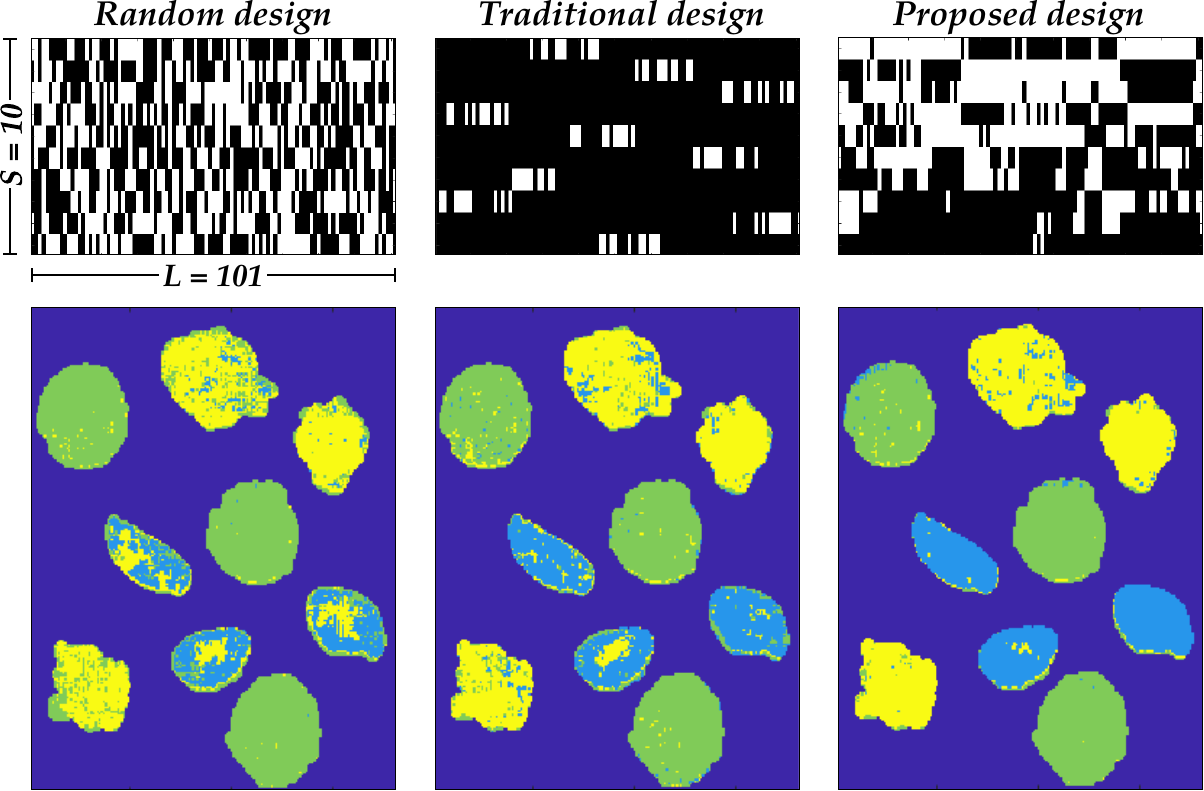}
    \caption{Visual representation of the classification results and coding patterns for each design using the $\%_{3D}=0.1$ scenario in the Cereals Dataset.}
    \label{fig:results_cereal}
\end{figure}

\begin{figure}[!t]
    \centering
    \includegraphics[width=1\columnwidth]{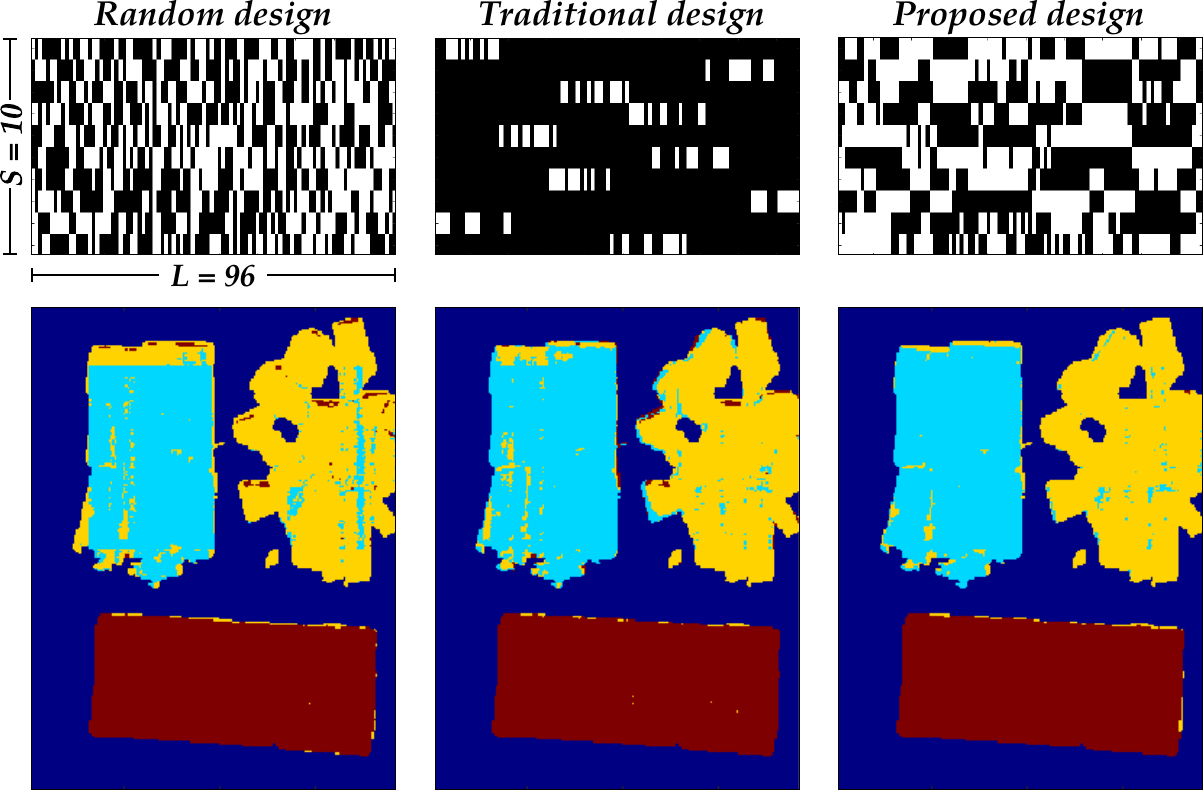}
    \caption{Visual representation of the classification results and coding patterns for each design using the $\%_{3D}=0.1$ scenario in the Yatsuhashi Dataset.}
    \label{fig:results_yatsu}
\end{figure}

Finally, in order to see a visual representation of the compressive NIR spectral classification, the bottom part of Figs. \ref{fig:results_cereal} and \ref{fig:results_yatsu} shows the results obtained for $\%_{3D}=0.1$. It can be seen that the proposed method is more accurate in the classification task for both datasets. Also, in the upper part of Figs. \ref{fig:results_cereal} and \ref{fig:results_yatsu}, the coding patterns are shown. The proposed Deep-design converges to special bandpass filters, similar to the traditional approach. However, the Deep-design contains more elements in one resulting in the optimal transmittance for the NIR dataset used.

\section{Conclusions}
This work presented a binary coding patterns design for compressive NIR spectral classification through modeling the 3D-CASSI system as an optical layer in an end-to-end model. To demonstrate the capabilities of the approach, it was successfully applied on two NIR spectral data cubes and compared with the Random-design, Traditional-design, and Full-data (without compression). In general, the results show that performing the classification directly on the compressive measurements ($\%_{3D}=0.5$) for the proposed design (Deep-design) provides similar accuracy results, compared with those provided by performing the classification on the Full-data. Additionally, based on the executed experiments using different sensing ratios, the Deep-design outperforms Traditional-design and Random-design for all compression scenarios. Also, in the extreme case of compression ($\%_{3D}=0.1$), the proposed design outperforms the other designs by up to $10\%$, which demonstrated the effectiveness of the optimized sensing matrix.

% -------------------------------------------------------------------------
\bibliographystyle{IEEEbib}
\footnotesize
\bibliography{main}

\end{document}